\RequirePackage{fix-cm}
\documentclass[smallextended]{svjour3}       % onecolumn (second format)
\smartqed  % flush right qed marks, e.g. at end of proof
\usepackage{url,graphicx,graphics,amssymb,amsmath,relsize,supertabular}
\usepackage[pdfstartview={FitH},pdffitwindow=true,colorlinks=true,citecolor=blue,linkcolor=blue]{hyperref}%
\usepackage{array,longtable}

\usepackage{mathptmx}      % use Times fonts if available on your TeX system
\usepackage{latexsym}
\begin{document}

\title{Investigation of commuting Hamiltonian in quantum Markov network%\thanks{Grants or other notes
}


\author{Farzad Ghafari Jouneghani, \and
        Mohammad Babazadeh \and
        Rogayeh Bayramzadeh \and
        Hossein Movla %etc.
}


\institute{Farzad Ghafari Jouneghani, Mohammad Babazadeh, Rogayeh Bayramzadeh, \at
              Department of Physics, University of Tabriz, Tabriz, Iran. \\
              Tel.: +989398990564\\
              \email{f.ghafari89@ms.tabrizu.ac.ir}           %  \\
           \and
          Hossein Movla, \at
              Azar Aytash Co., Technology Incubator, University of Tabriz, Tabriz, Iran
}

\date{Received: date / Accepted: date}

\maketitle

\begin{abstract}
Graphical Models have various applications in science and engineering which include physics, bioinformatics, telecommunication and etc. Usage of graphical models needs complex computations in order to evaluation of marginal functions, so there are some powerful methods including mean field approximation, belief propagation algorithm and etc. Quantum graphical models have been recently developed in context of quantum information and computation, and quantum statistical physics, which is possible by generalization of classical probability theory to quantum theory. The main goal of this paper is preparing a primary generalization of Markov network, as a type of graphical models, to quantum case and applying in quantum statistical physics. We have investigated the Markov network and the role of commuting Hamiltonian terms in conditional independence with simple examples of quantum statistical physics.
\end{abstract}

\section{Introduction}
\label{intro}
In 1988, Pearl devised belief propagation algorithm to solve marginalization and other inference problems. Belief propagation algorithm is a message-passing algorithm using graphs display known as graphical models to solve the problems \cite{Pearl1}.On the other side, Gallager, in 1963, devised a decoding algorithm similar to the belief propagation algorithm to decode his well-known and valuable low-density-parity-check code. One of the significant points of graphical models' capability and generalization is that they appear in variant field of science dealing with mathematics including code theories \cite{Gallager2,McEliece3},physics \cite{MacKay4,Yedidia5}, statistics, and artificial intelligence \cite{Pearl1,Kschischang7}. This diversity, unfortunately, has caused these models and related algorithms lack a single and standard notation. Numerous generalizations, from initial message-passing algorithm to highly complicated approaches, have been made about belief propagation algorithm.Generalization of graphical models and related algorithms on quantum mechanics has been recently noticed by researchers of this field. Poulin et al. worked on this field in 2008. Afterwards, its application was investigated in quantum statistical mechanics \cite{Poulin9,Hasting10,Bilgin11}, and quantum codes \cite{Pawel12}and there were acceptable results. One of the main reasons for this generalization is the quantum mechanics nature as a probabilistic theory being originally identical to graphical models. Despite the basic similarity, the differences between quantum mechanics and classic probability theory have complicated the generalization. Therefore, it can be highly challenging to structure an applicable and unique framework to which we can apply both quantum mechanics and graphical models simultaneously.

At the beginning of this paper, theoretical foundation of classic and quantum Markov network were explained in graph language and their relation with quantum and statistical physics was expanded as well. We have already tried to have an overall review on previous researches, so we hope this paper is an opening to familiarizing with graphical models and it may have interesting ideas about quantum.\\
\section{Pairwise Markov Random Field (MRF)}
\label{sec:1}
Pairwise Markov random field was used as an inference model in computer vision early last decade \cite{Hovington13}. It may seem indisputable to some computer science researchers to solve problems like objects recognition or other preliminary computer vision problems. What we are looking for in graphical models is to give some data and get the answer. In computer vision, for instance, we give a series of two-dimensional pixels and expect to receive optimal image. This is challenging by itself.
\\To solve computer vision, MRF has applied a highly interesting theoretical model \cite{William14}. What we are generally looking for in these problems is to guess the input in terms of output. To elucidate more, suppose that we have a $100\times100$ black and white output image sent to us through a channel. Some of these pixels are maybe flipped or their colors are changed (black to white and white to black) while transferring. We intend to reconstruct the original image (as it was before transferring through the channel).
Generally, we aim to infer input data \ensuremath{x_i} out of output data \ensuremath{y_i} (each pixel variable). Index \ensuremath{i} may indicate pixel's place or a small collection of pixels. Another assumption is that, there is a statistical relation between \ensuremath{x_i} and \ensuremath{y_i} for each \ensuremath{i} called compatibility function \ensuremath{\Phi_i(x_i,y_i)}.  Function \ensuremath{\Phi_i(x_i,y_i)} is usually called as an evidence of \ensuremath{x_i}. Eventually, to well infer \ensuremath{x_i} we have to consider a structure for it. Neglecting such a structure will lead incomplete computer vision problems. To understand the general shape of this structure, consider the \ensuremath{i}th variable in a two-dimensional square network. Variable \ensuremath{x_i} should be compatible with the nearest variable \ensuremath{x_j} which is shown by compatibility function \ensuremath{\Psi_{ij}(x_i,y_j)} where \ensuremath{\Psi_{ij}} only connects neighboring points; the total probability function is as follows:
\begin{equation}
p(\{x\},\{y\})=\frac{1}{z}\prod_{ij}\Psi_{ij}(x_i,y_j)\prod_{i}\Phi_i(x_i,y_i)
\end{equation}
\\Where Z is a normalization constant and the product on (i, j) consist all neighboring points in the square network. The above graphical model is shown in \cite{Geman15}. Filled circles indicate observed values \ensuremath{y_i} while unfilled circles show hidden values \ensuremath{x_i}. Since compatibility functions only depend on one pair of nodes or variables \ensuremath{i} and \ensuremath{j}, MRF was called pairwise.\\
\begin{figure}
%% Use the relevant command to insert your figure file.
%% For example, with the graphicx package use
\includegraphics[width=0.6\textwidth]{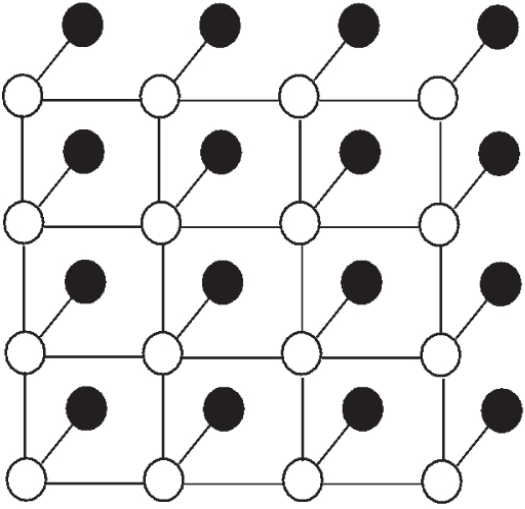}
\caption{two-dimensional square pairwise MRF network}
\label{fig:1}       % Give a unique label
\end{figure}

\section{MRF based on graphs}
\label{sec:2}
Graphical models having different applications serve in numerous areas including Bayesian networks(artificial intelligence), factor graph(image reconstruction), Tanner graph(code theory), and Markov networks(statistical physics). Besides naming, these models bear numerous common attributes which are based on graphs. Graphical models are completely convertible to each others. In this section, hence, emphasizing MRF, we try to mathematically expand graphical models in the form of graph theories which has a significant role in graphical models generalization on quantum counterpart.
Each graphical model is shown by $G= (V, E)$, where V and E are collections of vertices and edges, respectively,and a graph vertex \ensuremath{x\in V} is given to each random variable x. In fact, each graphical model denotes a distribution as \ensuremath{P(V)=P(x_1,x_2,...)}and the edge $e= (u, v)$, $e\in E$ shows the dependence between random variables $u$ and $v$ in probability distribution $P$.\\\\
\textbf{Conditional independence:} Take $A$, $B$, and $C$ as three random variable sets wit distribution $P (A, B, C)$. $A$ and $C$ are independent given $B$ if conditional mutual information \ensuremath{I(A:C \mid B )=0}:
\begin{equation}
I(A:C \mid B)=H(A,B)+H(C,B)-H(A,B,C)-H(B)
\end{equation}\\
Where $H$ is Shannon entropy. And for each arbitrary distribution, $p(x)$ is defined as \cite{Nielsen16}:
\begin{equation}
H(P(x))=-\sum_{x}p(x)\ln p(x)
\end{equation}\\
\textbf{Markov Random Field (MRF):} Graph $G= (V, E)$ and distribution $P(V)$ are given. Pair $(G,P(V))$ creates the MRF If \ensuremath{I(U:V-U-n(U)\mid n(U)=0}) for each $U \subset V$ where $n(U)$ is the neighboring sets of $U$. In other words, $U$ is protected by neighboring points \cite{Jonathan17}(fig 2 \cite{Lauritzen18}).
\begin{figure}
%% Use the relevant command to insert your figure file.
%% For example, with the graphicx package use
\includegraphics[width=0.6\textwidth]{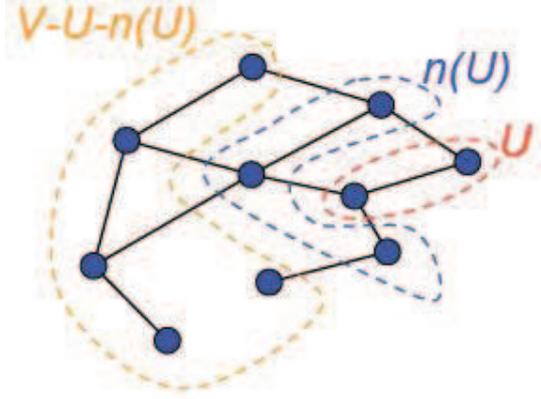}
\caption{Markov Random Field (MRF)}
\label{fig:2}       % Give a unique label
\end{figure}\\
\textbf{Clique:}is a set of graph nodes which are all connected together. The set of all cliques of graph G is shown by C(G). Cliques are shown in Figure 3.\\
\begin{figure}
\includegraphics[width=0.6\textwidth]{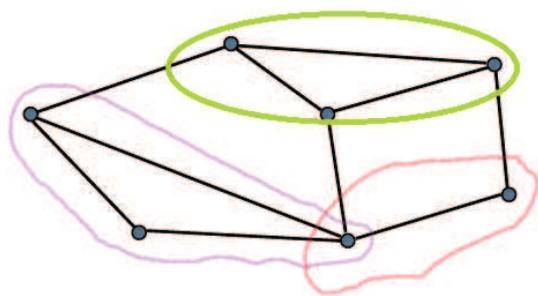}
\caption{illustration of cliques which belong to MRF}
\label{fig:3}
\end{figure}\\
\textbf{Hammersley Clifford theorem \cite{Matt19}:}Pair $(G, P(V))$ creates a positive MRF $(P>0)$ if and only if the distribution function can be written in following form:
\begin{equation}
P(V)=\frac{1}{Z}\prod_{c \in C(G)}W(c)
\end{equation}
$W(c)$ is the local probability distribution function on cliques and $Z$ is the normalization constant. See \cite{Hammersley20} to familiarize more with graphical models.\\
\textbf{Example}: Pairwise MRF: When the maximum number of available nodes in cliques is 2, then MRF distribution can be written as:
\begin{equation}
P(V)=\frac{1}{Z}\prod_{x \in V}W(x)\prod_{(x,y)\in E}\mu(x,y)
\end{equation}
Looking more precisely at above definitions, we can infer a highly-rich relation between them. Conditional independence which gives a highly innovative idea of belief propagation algorithm, plays a basic role in MRF definition. To make a background, we pose a discussion here.\\
\begin{figure}
\includegraphics[width=0.6\textwidth]{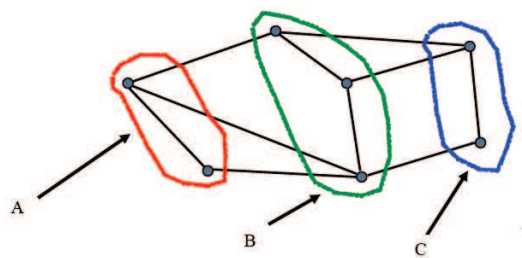}
\caption{MRF with three set of nodes: $A, B$ and $C$}
\label{fig:4}       % Give a unique label
\end{figure}
Suppose that the sets A and C in figure 4 \cite{D.Poulin21} are independent given B. If we want to determine the probability of B being in a certain state, we should consider its interaction with other nodes of graph. We also know that if we determine B, the relation between A and C is totally disconnected. So we can separately investigate the interaction of "B and A", and "B and C" which is of significant help in decreasing calculation complexities. It can be said:\\\\
\textbf{The probability of B in a certain state= $P(B \mid C) \times P(B \mid A)$}\\\\
The above explanations show the importance of conditional independence in defining MRF. However, we know that calculating conditional mutual information for all complicated arbitrary nodes in large graphs is time consuming and practically impossible. Hammersley-Clifford theorem is a shortcut to avoid such calculations. By a graph's shape or its corresponding distribution, this theorem shows that whether or not it is MRF. So, by connecting Hammersley-Clifford theorem and conditional independence property in defining MRF, we can access conditions in which calculation complexity is highly reduced.
\section{Quantum Markov Network}
Now we want to generalize classical concepts of graphical models on quantum. Before doing so, it is worth noting that this generalization is used in quantum mechanics and statistical physics, and interested readers can see \cite{Leifer8}. Each graphical model is classically shown by graph $G= (V, E)$ where $V$ shows nodes and $E$ shows lines called edges. Variables \ensuremath{x_i}, here, are quantum systems e.g. spin and local functions operate in Hilbert space. General distribution function is density matrix,\ensuremath{\rho} , which is as below for a pairwise MRF:
\begin{equation}
\rho=\prod_{x \in V}v_{x}\prod_{(x,y)\in E}\mu_xy
\end{equation}
\\In above equation, $v_{x}$ and $\mu_xy$ are positive operators in Hilbert space $H_{x}$ and $H_{x}\otimes H_{y}$ respectively. $\rho$ operates in Hilbert space $H=\otimes_{x \in V}H_x$ and the edge $e=(x,y)$ shows the dependency between variables $x$ and $y$ in $\rho v$. We work with Gibbs distribution in statistical physics  which is $\rho \equiv \frac{1}{Z} \exp(- \beta H)$, where $H$ and $\beta$ are Hamiltonian and inverse of temperature, respectively.\\
\textbf{Conditional independence:}If we take $A$, $B$, and $C$ as three quantum systems with joint distribution $\rho(A, B, C)$, then $A$ and $C$ are independent given B if conditional mutual information $I(A:C\mid B)$ =0.
\begin{equation}
I(A:C\mid B)=S(A,B)+S(C,B)-S(A,B,C)-S(B)
\end{equation}\\
$S$ is Von-Neumann entropy and for each arbitrary density matrix is defined as:
\begin{equation}
S(\rho)=-tr(\rho\ln(\rho))
\end{equation}\\
\textbf{Quantum Markov Network\cite{Leifer8,Brown22}:}Graph $G=(V, E)$ and density matrix $\rho$ are given. Pair $(G,\rho (V))$ creates MRF if $I(A:C\mid B)=0$ for each arbitrary $A$, $B$, and $C$, where $A$, $B$, and $C$ are separate subsets of $V$ so that $B$ protects $A$ against $C$. On the other hand, $B$ is the collection of neighboring points of $A$ and $C=V-A-B$ (figure 5 \cite{D.Poulin21}).
\begin{figure}
\includegraphics[width=0.6\textwidth]{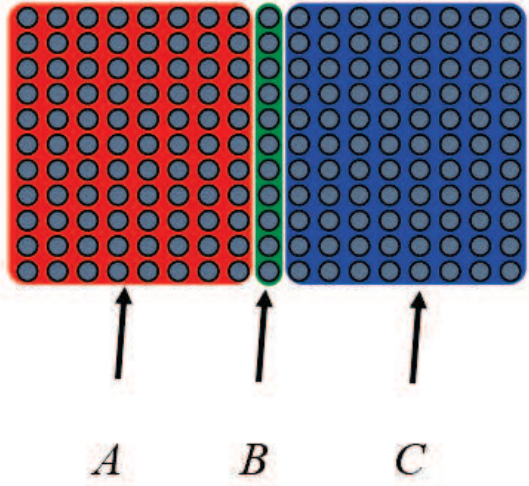}
\caption{$A,B$ and $C$ for an arbitrary MRF}
\label{fig:5}       % Give a unique label
\end{figure}
\\So far, only classical concepts explained before, have been generalized. However, in quantum systems, we encounter non-commutativity of operators. So, we should be a bit cautious in generalization of MRF to quantum, because in quantum counterpart, it is not easy to understand conditional independency through graph shape or corresponding distribution. We will come to this challenge later. to elucidate the topic, we take a look at an example from statistical physics in which the system is classic and H is defined as:
\begin{subequations}
\begin{equation}
H(x_{1},x_{2},x_{3},...)=h(x_{1},x_{2})+h(x_{2},x_{3})+...
\end{equation}
\begin{equation}
Z=\sum_{x_1,x_2,...} e^{-H(x_1,x_2,...)}       \text{Partition Function}
\end{equation}
\begin{equation}
=\sum_{x_2,x_3,...}\underbrace{ \lgroup \sum_{x_1}e^{-h(x_{1},x_{2})} \rgroup}_{Z_{1,2}(x_2)}
\end{equation}
\begin{equation}
=\sum_{x_3,x_4,...}\underbrace{ \lgroup \sum_{x_2}Z_{1,2}(x_2)e^{-h(x_{2},x_{3})} \rgroup}_{Z_{2,3}(x_3)}
\end{equation}
\begin{equation}
=\sum_{x_N}Z_{N-1,N}(x_N)e^{-h(x_{N-1},x_{N})}
\end{equation}
\end{subequations}
Now we are ready to generalize these equations to their quantum counterpart:
\begin{subequations}
\begin{equation}
H(x_{A},x_{B},x_{C},...)=h(x_{A},x_{B})+h(x_{B},x_{C})+...
\end{equation}
\begin{equation}
h_{BC}=I_{A}\otimes h_{BC}\otimes I_{D}\otimes ... \textrm{Two-body hermitian operator}
\end{equation}
\begin{equation}
Z=Tr(e^{-H(x_{A},x_{B},x_{C},...)})
\end{equation}
\begin{equation}
\neq Tr_{BC...}\{ Tr_{A}(e^{-h(x_{A},x_{B})})e^{-h(x_{x_{B},x_{C}})...}\}
\end{equation}
\end{subequations}\\
equality does not happens because Hamiltonian terms dose not commute.\\\\
\textbf{Quantum Hammersley Clifford Theorem:\cite{Leifer8,Brown22}}
Now the question arises is that if we can generalize Hammersley Clifford theorem to quantum Or more precisely, if this theorem can absolutely identify whether a quantum statistical system with Gibbs distribution is MRF or not? In reply we can say that there is a quantum counterpart for Hammersley Clifford theorem. The only difference is that, in quantum, we encounter the above limitations, And this theorem in quantum state is only true in one direction.\\\\
\textbf{Theorem 1}\cite{Leifer8}: Let $G =(V,E)$ be a graph and $\rho$ be a density matrix for the particles located at the vertices of G. If the pair ($\rho$,G) is a positive quantum Markov network, then the state $\rho$ can be expressed as $\rho=e^H$ where $H= \sum_{Q\in C} h_Q$ is the sum of Hermitian operators $h_{Q}$ on the particles located in cliques Q.\\
For the reverse direction, we have can consider the below theorem.\\\\
\textbf{Theorem 2}\cite{Brown22}: Let $G=(E,V )$ be a graph and $H= \sum_{Q\in C} h_Q$ , $[h_{Q},h_{Q'}]=0$ be a local commuting Hamiltonian on that graph. Then ($\rho$,G) is a positive quantum Markov network for $\rho=\frac{1}{z}e^H$ ,where $Z$ is a normalization constant.\\\\
\textbf{The reverse happens when all $h_{Q}$s commutate together.}\\\\
\textbf{Example:} It is the right time for us to investigate these claims in just simple examples. Our example is a system with five spins on which applied local magnetic fields. The following Hamiltonian describe this system, while $S^{i}$ are Pauli matrixes (see figure 6).
\begin{subequations}
\begin{equation}
h_{12}=(S^{x}\otimes S^{x})\otimes I\otimes I\otimes I+h_{1}(S^{z}\otimes I \otimes I\otimes I\otimes I)
\end{equation}
\begin{equation}
h_{23}=I\otimes(S^{x}\otimes S^{x})\otimes I\otimes I+\frac{1}{2}h_{2}(I\otimes I\otimes S^{z}\otimes I\otimes I)
\end{equation}
\begin{equation}
h_{34}=I\otimes I\otimes(S^{x}\otimes S^{x})\otimes I+\frac{1}{2}h_{2}(I\otimes I \otimes S^{z}\otimes I\otimes I)
\end{equation}
\begin{equation}
h_{45}=I\otimes I\otimes I\otimes (S^{x}\otimes S^{x})+h_{3}(I\otimes I\otimes I\otimes I\otimes S^{z})
\end{equation}
\begin{equation}
H=h_{12}+h_{23}+h_{34}+h_{45}
\end{equation}
\end{subequations}
Where h's describe local magnetic fields. What we anticipate is that conditional $I$ over different parts of relevant graph must be nonzero, because there are some non-commutative terms in Hamiltonian. For a sample, we can see figure 7 which illustrate the $I(A:C\mid B)$, where $A$, $C$ and $B$ comprise spins $(1 \& 2)$, $(4 \& 5)$ and $3$ respectively. Putting $h_{1},h_{2},h_{3}=2$, in the figure 7 as we can see $I(A:C\mid B)$ is nonzero except for $\beta=0$, because $h_{ij}$’s for $i,j=1,…,5$ are not commuting.
Now what happens if we put $h_{2}=0$? AS a result of this manipulation we have all $h_{ij}$’s commuting. Consequently, as we can see in figure 8 for $I(A:C\mid B)$ described above is zero for all $\beta$. These examples make an intuitive sense for the Quantum Hammersley-Clifford theorem.
\begin{figure}
\includegraphics[width=0.6\textwidth]{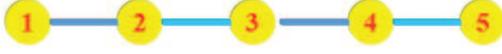}
\caption{a system with five spins}
\label{fig:6}
\end{figure}
\begin{figure}
\includegraphics[width=0.6\textwidth]{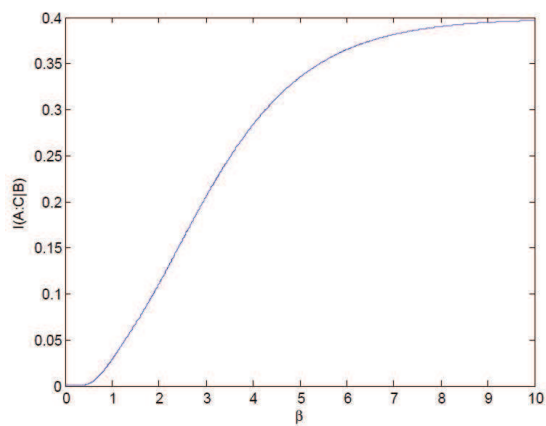}
\caption{$I(A:C\mid B)$ for a five-particle system, where $A, B$ and $C$ comprise spins $(1 \& 2), (4 \& 5)$ and 3 respectively and  $h_{1},h_{2},h_{3}=2$}
\label{fig:7}
\end{figure}\\
\begin{figure}
\includegraphics[width=0.6\textwidth]{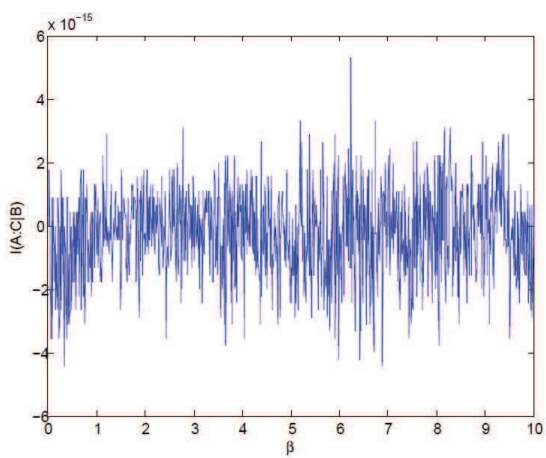}
\caption{$I(A:C\mid B)$ for a five-particle system, where $A$, $B$ and $C$ comprise spins $(1 \& 2)$, $(4 \& 5)$ and $3$ respectively, with $h_{1}$ ,$h_{3}=2$ and $h_{2}=0$}
\label{fig:8}
\end{figure}\\
\section{Conclusion}
The authors merely aimed at posing primary ideas and questions. In this paper, it is primarily guessed that commutativity equals being Markov in quantum case. This guess can be highly valuable for if we can identify quantum Markov network, then we can use conditional independence and so inference problems in quantum mechanics will be much easier. Now, we are in a state to use message-passing algorithm in graphical models for inference problems in quantum mechanics. This proposition by itself can form into a new research. On the other side, graphical models based on quantum physics form a problem which can attract enthusiasts.
In the following, there are some recommendations for those who are interested in. One of them is more investigation on quantum graphical models and generalization of MRF to quantum. More precisely, when can we say strongly that a quantum distribution and its related graph is a quantum MRF? On the other side, can we answer this question decisively? Or only under certain conditions we are able to say whether or not a quantum distribution is MRF.

\end{document}